\documentclass[twocolumn]{IEEEtran} 
\usepackage[T1]{fontenc}
\usepackage{color}
\usepackage{mathrsfs}
\usepackage{amsmath}
\usepackage{amsthm} 
\usepackage{amssymb}
\usepackage{graphicx}
\usepackage{cite}
\usepackage{makecell}
\usepackage{algorithm}
\usepackage{algorithmic}
\usepackage{fancyhdr}
\usepackage{diagbox}
\usepackage{bm}
\usepackage[unicode=true,
 bookmarks=true,bookmarksnumbered=true,bookmarksopen=true,bookmarksopenlevel=1,
 breaklinks=false,pdfborder={0 0 0},pdfborderstyle={},backref=false,colorlinks=false]
 {hyperref}
\hypersetup{
 pdftitle={Your Title},
 pdfauthor={Your Name},
 pdfpagelayout=OneColumn, pdfnewwindow=true, pdfstartview=XYZ, plainpages=false}

\usepackage{multirow} 
\usepackage{xcolor}

\allowdisplaybreaks[4]

\usepackage[caption=false,font=footnotesize]{subfig}

\IEEEoverridecommandlockouts

\usepackage{geometry}
\begin{document}

\title{\textcolor{black}{Prompt-Driven Low-Altitude Edge Intelligence: Modular Agents and Generative Reasoning}

\author{\IEEEauthorblockN{Jiahao You, Ziye Jia, \textit{Member, IEEE,} Chao Dong, \IEEEmembership{Senior Member,~IEEE,} and Qihui Wu, \IEEEmembership{Fellow,~IEEE}}
\thanks{Jiahao You, Chao Dong, and Qihui Wu are with the Key Laboratory of Dynamic Cognitive System of Electromagnetic Spectrum Space, Nanjing University of Aeronautics and Astronautics, Nanjing 211106, China (e-mail: yjiahao@nuaa.edu.cn, dch@nuaa.edu.cn, wuqihui@nuaa.edu.cn).

Ziye Jia is with the Key Laboratory of Dynamic Cognitive System of Electromagnetic Spectrum Space, Nanjing University of Aeronautics and Astronautics, Nanjing 211106, China, and also with the National Mobile Communications Research Laboratory, Southeast University, Nanjing, Jiangsu, 211111, China (e-mail: jiaziye@nuaa.edu.cn).}}}

\maketitle

\begin{abstract}
The large artificial intelligence models (LAMs) show strong capabilities in perception, reasoning, and multi-modal understanding, and can enable advanced capabilities in low-altitude edge intelligence. 
However, the deployment of LAMs at the edge remains constrained by some fundamental limitations. 
First, tasks are rigidly tied to specific models, limiting the flexibility. Besides, the computational and memory demands of full-scale LAMs exceed the capacity of most edge devices. Moreover, the current inference pipelines are typically static, making it difficult to respond to real-time changes of tasks.
To address these challenges, we propose a prompt-to-agent edge cognition framework (P2AECF), enabling the flexible, efficient, and adaptive edge intelligence. 
Specifically, P2AECF transforms high-level semantic prompts into executable reasoning workflows through three key mechanisms. 
First, the prompt-defined cognition parses task intent into abstract and model-agnostic representations. 
Second, the agent-based modular execution instantiates these tasks using lightweight and reusable cognitive agents dynamically selected based on current resource conditions. 
Third, the diffusion-controlled inference planning adaptively constructs and refines execution strategies by incorporating runtime feedback and system context.
In addition, we illustrate the framework through a representative low-altitude intelligent network use case, showing its ability to deliver adaptive, modular, and scalable edge intelligence for real-time low-altitude aerial collaborations.
\end{abstract}


\section{Introduction}\label{s1}
The large artificial intelligence models (LAMs), such as generative pre-trained transformer 4 (GPT-4), Gemini, and segment anything model (SAM), significantly advance the frontiers of perception, reasoning, and multi-modal understanding \cite{AIGC_2024,LAM_2025}. Enabled by deep neural architectures, extensive parameterization, and training on diverse datasets, these models demonstrate exceptional capabilities across complex cognitive tasks, including language interpretation, visual recognition, and intelligent decision-making. However, the deployment of LAMs predominantly occurs in centralized cloud environments, where abundant computational resources support their massive parameter scales. Despite the strong performance, the cloud-based LAM deployment inherently suffers from high latency, substantial bandwidth consumption, and potential reliability concerns, which limits their applicabilities in latency-sensitive and mission-critical applications \cite{VTM_2024}.

\begin{figure*}[!t]
\centering
\includegraphics[width=16cm]{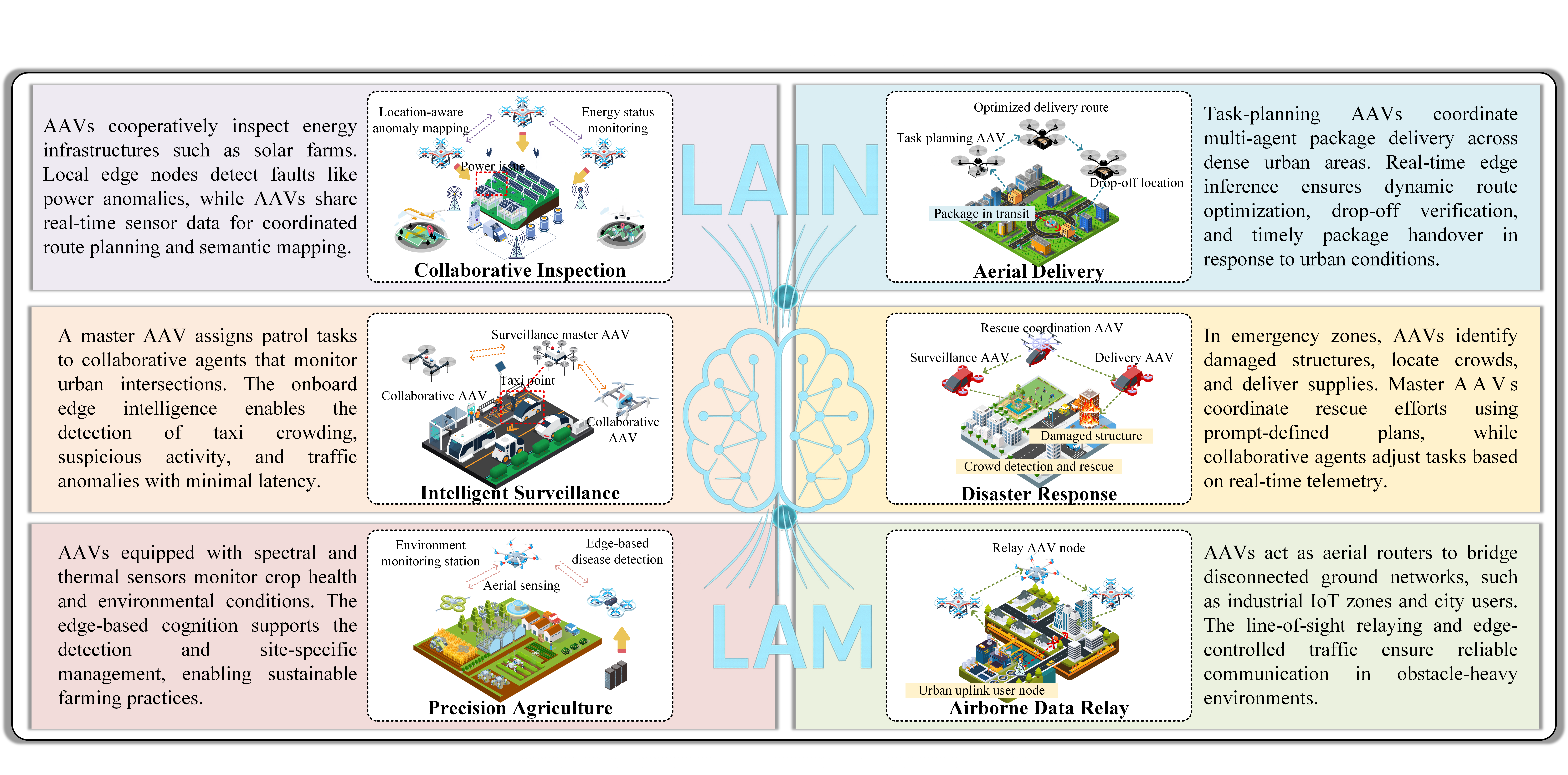}
\caption{Representative LAIN application scenarios showing AAV-edge cooperation for collaborative inspection, aerial delivery, intelligent surveillance, disaster response, precision agriculture, and airborne data relay.}
\label{f1}
\end{figure*}

At the same time, the low-altitude intelligent networks (LAINs) emerge as a promising paradigm, designed explicitly to embed the cognition directly within distributed low-altitude infrastructures. The architectures of LAIN integrate airborne platforms, such as autonomous aerial vehicles (AAVs), and terrestrial edge nodes to form dynamic, context-aware communication and computation ecosystems. 
As illustrated in Fig. \ref{f1}, the LAINs are particularly suited for real-time, autonomous, and collaborative applications, including the aerial surveillance, intelligent transportation, urban emergency response, and disaster management \cite{Security_2023,jia2025}. By shifting the intelligence closer to users and data sources, LAINs enable rapid, localized decision-making and adaptive responses, which are critical for the dynamic low-altitude missions.

Nevertheless, deploying LAMs within LAIN presents fundamental architectural and operational challenges. First, most existing edge intelligence frameworks rely on rigid task-to-model bindings, where each cognitive task is statically linked to specific models, which severely restricts the flexibility and adaptability to evolving missions \cite{Elastic_2024,Joint_You_2025}. 
 Second, the computational and memory demands associated with LAMs, which often reach billions of parameters, significantly exceed the capacity of resource-constrained edge nodes, even when applying compression techniques such as pruning and quantization \cite{UAV_USV_2024}.
Finally, the traditional edge inference workflows are typically static and context-agnostic, and they fail to dynamically adjust their reasoning processes in response to fluctuating environmental conditions, user intent, or real-time system status \cite{Diffusion_Based_2024}.

Recent studies explore various strategies to alleviate the architectural and operational challenges of deploying LAMs in edge environments. Kim \textit{et al.} \cite{LLM_2025_Compressing} provide a taxonomy of model compression and parameter-efficient fine-tuning methods, including quantization, pruning, knowledge distillation, and low-rank adaptation, which aim to reduce the model size and computational cost while maintaining performance.
Zheng \textit{et al.} \cite{LLM_2025_Design} present a holistic framework for edge-optimized LAM deployment, addressing model design, scheduling, and runtime adaptation for heterogeneous hardware. Liu \textit{et al.} \cite{PR_LLM_2025_Design} explore prompt-based learning as a flexible interface for task specification, enabling zero-shot or few-shot reasoning without task-specific retraining.
However, the existing approaches lack supports for semantic abstraction, adaptive execution, and system-level integration. They often treat the edge inference as a static, scaled-down version of cloud models, which limits the flexibility and responsiveness in dynamic environments.

To address these challenges, this article proposes a prompt-to-agent edge cognition framework (P2AECF). The framework integrates three complementary mechanisms: prompt-defined cognition, agent-based modular execution, and diffusion-controlled inference planning. Within P2AECF, high-level semantic prompts provide a flexible and expressive interface for specifying cognitive tasks, which are then systematically decomposed into executable reasoning workflows. These workflows leverage lightweight and reusable cognitive agents dynamically assigned based on resource constraints and mission context. Moreover, a diffusion-inspired inference planner adaptively constructs and refines reasoning paths by continuously incorporating real-time feedback, contextual embeddings, and historical performance metrics.
The main contributions can be summarized in the following.

\begin{itemize}
  \item We present P2AECF, an edge cognition framework that converts high-level semantic prompts into executable reasoning workflows through the task graph abstraction. It decouples the cognition from fixed models and supports closed-loop, context-aware decision-making in LAIN.
  \item We develop a lightweight and metadata-driven agent framework that enables fine-grained cognitive functions to be dynamically instantiated, scheduled, and reused across heterogeneous nodes in LAIN. This design supports scalable and efficient edge reasoning under varying system capabilities and demands.
  \item We propose a diffusion-inspired scheduler that adaptively generates reasoning paths using prompt semantics, runtime signals, and historical performance, thereby improving the robustness and enabling continual optimization of edge cognition.
\end{itemize}

\begin{figure*}[!t]
\centering
\includegraphics[width=16cm]{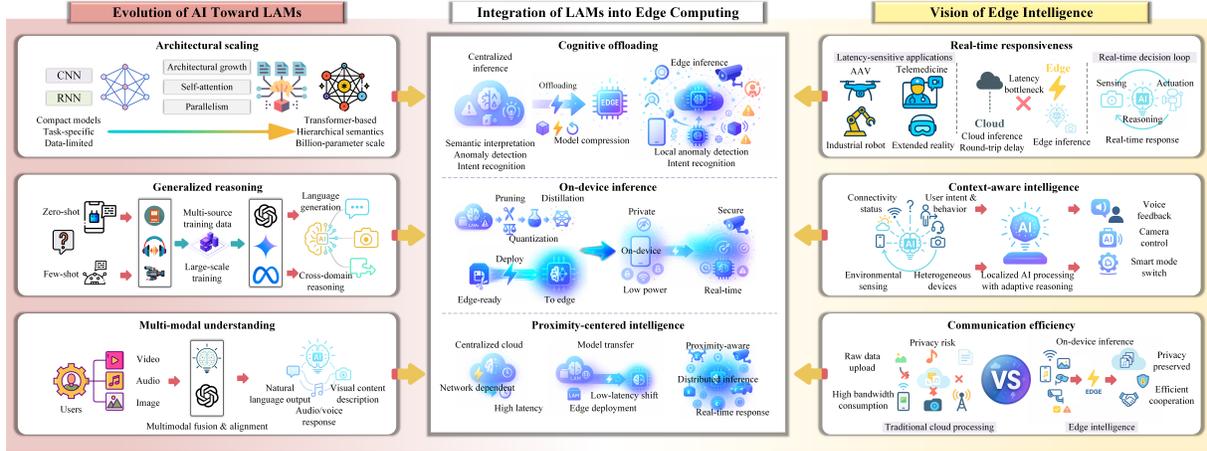}
\caption{The path to cognition-native edge intelligence: from LAMs to LAIN edge deployment, integrating cognitive offloading and on-device inference to enable real time, context aware operation.}
\label{f2}
\end{figure*}

\section{Cognition-Native Edge Intelligence}\label{s2}
The evolution of LAMs, emergence of edge-native intelligence in LAINs, and convergence as a pathway toward cognition enabled edge systems, are illustrated in Fig. \ref{f2}.

AI evolves from rule-based and statistical models to large-scale, general-purpose learning frameworks, marked by substantial improvements in architecture, scalability, and generalization.

\begin{itemize}
\item \textbf{Architectural scaling}: LAMs leverage deep transformer-based architectures with billions of parameters and self-attention mechanisms, enabling scalable and domain-agnostic intelligence \cite{Transformers_2022}.
\item \textbf{Generalized reasoning}: Models like GPT-4, Gemini, and SAM support zero-shot or few-shot learning across tasks without retraining, providing high flexibility and deployment efficiency \cite{MALLM_2024}.
\item \textbf{Multi-modal understanding}: LAMs unify vision, language, audio, and symbolic inputs within a shared semantic space, allowing rich and human-like interactions with diverse data \cite{MALLM_2024}.
\end{itemize}

LAINs shift the intelligence from the cloud to distributed edge nodes, enabling low-latency and context-aware cognition aligned with real-time environmental dynamics.

\begin{itemize}
\item \textbf{Real-time responsiveness}: Applications like AAV coordination and extended reality demand sub-millisecond latency. The on-device decision-making avoids cloud delays and ensures reliable control \cite{ULL_2021}.
\item \textbf{Context-aware intelligence}: LAINs adapt to dynamic conditions and user intent using localized sensor inputs, without the constant cloud dependence.
\item \textbf{Communication efficiency}: Processing data at the edge reduces bandwidth, preserves privacy, and supports scalable multi-agent collaboration via semantic-level information exchange.
\end{itemize}

Bringing LAMs to the edge is essential for enabling cognition-native services in LAINs. This integration unfolds through three key trends:

\begin{itemize}
\item \textbf{Cognitive offloading}: Edge nodes increasingly handle tasks like semantic interpretation and anomaly detection using lightweight micro-models, reducing reliance on cloud infrastructure.
\item \textbf{On-device inference}: Advances in compression techniques and edge hardware (e.g., edge graphics processing units) enable efficient executions of LAM fragments directly on embedded platforms.
\item \textbf{Proximity-centered intelligence}: Locally deploying LAMs near data sources can improve latency, resilience, and context-awareness, supporting the personalized and mission-critical decision making.
\end{itemize}

\begin{figure*}[!t]
\centering
\includegraphics[width=15.5cm]{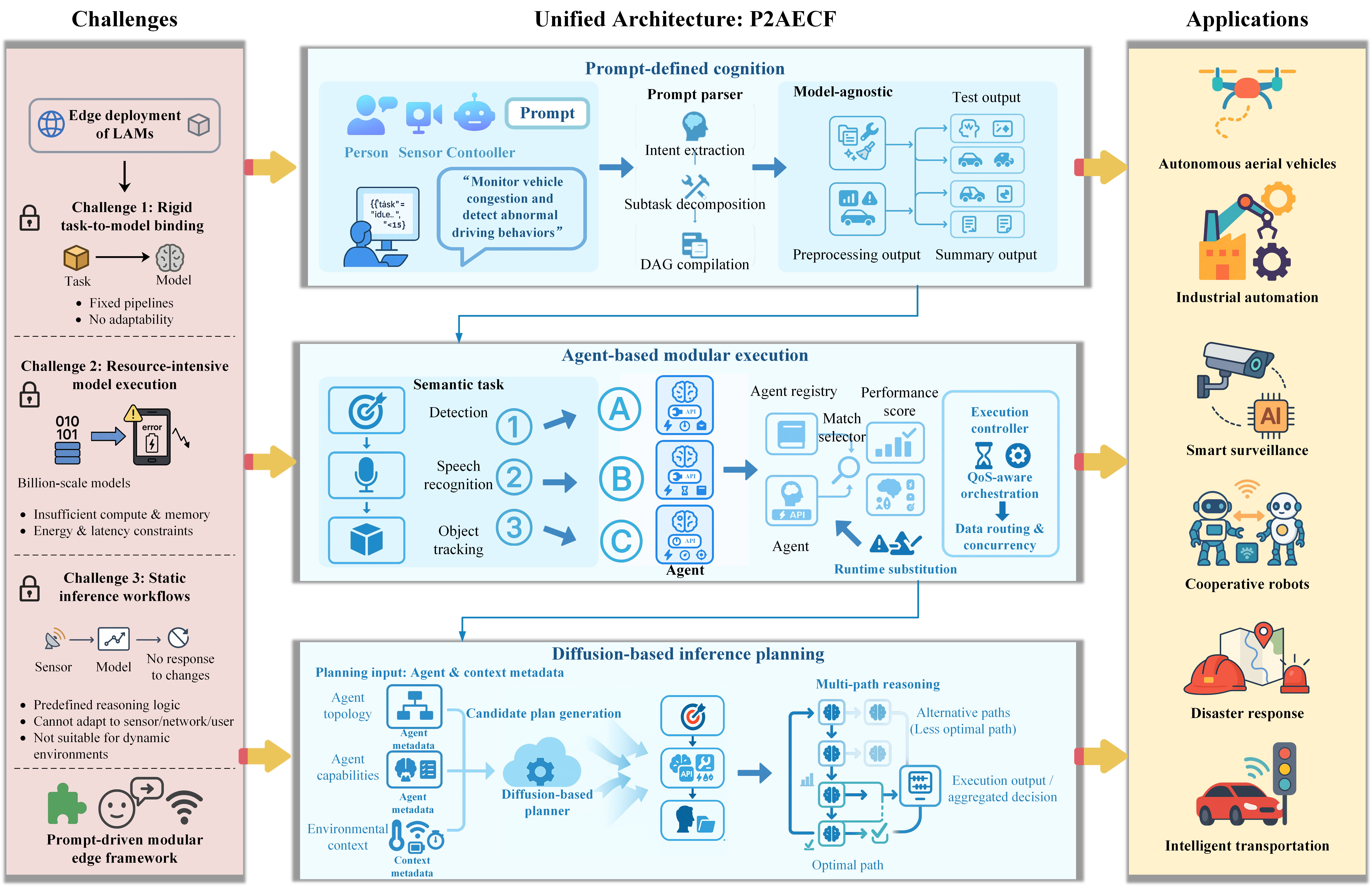}
\caption{Challenges, unified architecture P2AECF, and applications.}
\label{f3}
\end{figure*}

\section{Unified Architecture: P2AECF}\label{s3}
The proposed P2AECF provides a unified and modular approach to enable edge-native cognitive reasoning for LAIN applications. 
This architecture integrates three core mechanisms, namely prompt-defined cognition, agent-based modular execution, and diffusion-based inference planning, into a dynamic system that supports semantic-level task expression, modular model composition, and context-adaptive inference optimization. 
These components enable edge devices to construct and execute cognitive workflows on demand, allowing for flexible, interpretable, and self-evolving decision-making processes, as illustrated in Fig. \ref{f3}.

\subsection{Challenges of Deploying LAMs in LAINs}
Deploying LAMs at the edge presents three fundamental challenges that motivate the design of the proposed framework.

\begin{itemize}
  \item \textbf{Rigid task-to-model binding:} Most existing edge intelligence systems bind each cognitive task to a specific pre-trained model or fixed pipeline. This design limits flexibilities and prevents edge nodes from adapting to new or evolving tasks without retraining or manual reconfiguration.  
  \item \textbf{Resource-intensive model execution:} The state-of-the-art LAMs contain hundreds of millions to billions of parameters, and demand substantial memory, compute power, and energy. Even with compression techniques such as pruning or distillation, edge devices often lack the capacity to support real-time inference under strict latency or energy constraints.  
  \item \textbf{Static inference workflows:} The current edge inference pipelines are predefined and remain agnostic to runtime variations in sensor inputs, network conditions, or user intent. As a result, they cannot reconfigure reasoning strategies dynamically, which is critical in highly dynamic environments such as multi-agent coordination or emergency response.
\end{itemize}

\subsection{Prompt-defined Cognition}
As the semantic entry point of the P2AECF pipeline, this module plays a crucial role in bridging human or system intent with downstream executable reasoning structures. It enables users, sensors, or higher-layer controllers to define tasks in an interpretable and declarative format.

These prompts can be expressed in the natural language (e.g., monitor vehicle congestion and detect abnormal driving behaviors), or structured formats such as javascript object notation (JSON) templates and domain-specific graphs. For instance, an operator may issue a prompt via the voice or dashboard input, while automated agents may generate structured prompts based on sensor triggers or scheduled missions. This flexibility supports both human-in-the-loop control and full automation.

Upon receiving a prompt, the framework invokes a parser that performs multi-stage semantic analysis. First, it extracts the core task intent, input-output bindings, operational constraints, and target objectives. Second, it segments the prompt into logical subtasks (e.g., data preprocessing, detection, and summarization), identifying potential parallelism and dependencies. Third, it compiles these elements into a formal representation in the form of a directed acyclic graph (DAG), where each node denotes an atomic cognitive operation.

These nodes remain model-agnostic, meaning they define "what" needs to be done without specifying "how." This graph-based abstraction not only decouples the task definition from the model implementation but also enables parallel execution and easy system evolution. It allows the same prompt to trigger different execution flows depending on context (e.g., available agents and network status), and supports reuse across applications. Additionally, prompts can embed meta-level constraints (e.g., response time bounds and energy budgets), enabling downstream modules to tailor plans accordingly.

\subsection{Agent-based Modular Execution}
As the core mechanism for task realization in the P2AECF framework, the agent-based modular execution bridges the gap between high-level semantic plans and resource-aware execution. This module enables flexible and scalable inference under edge constraints by dynamically composing reasoning pipelines from modular micro-models.

Once a prompt-defined task graph is compiled, each abstract node representing a semantic subtask such as object detection, anomaly recognition, or speech parsing is instantiated by assigning it to a lightweight and reusable cognitive agent.
These agents are pre-registered or automatically generated based on model libraries, and are designed to operate within the limited compute, memory, and energy budgets typical of edge environments.
Each cognitive agent is encapsulated with a standardized interface and a metadata profile, which includes descriptors such as supported input-output modalities (e.g., telemetry data), expected inference latency, energy consumption, and task accuracy on relevant benchmarks. This metadata supports the fine-grained matching between task requirements and system capabilities, and also facilitates the runtime optimization and cross-application reuse.
The agent selection process is both dynamic and context-aware. 
Given the system state, such as compute load, battery level, thermal constraints, or connectivity-the orchestrator maps each DAG node to the most suitable agent from the registry. 
This decision is further guided by historical performance logs, which provide insights into prior latency, reliability, and success rates under similar conditions.

Once deployed, agents are coordinated by a local execution controller, which manages data dependencies, schedules concurrent tasks, and ensures compliance with quality of service (QoS) requirements.
In case of agent failure or performance degradation, the controller supports runtime substitution, enabling the system to maintain operational resilience and adaptivity without full workflow reinitialization.

\subsection{Diffusion-based Inference Planning}

The diffusion-based inference planning serves as a generative reasoning module that replaces static, rule-based scheduling with a dynamic, context-aware mechanism tailored for complex edge environments. In the P2AECF architecture, this module is activated after the agent-level task graph has been instantiated, determining not only the execution order but also the conditional resource allocation and timing strategy for each cognitive agent.

The traditional inference pipelines rely on predefined or heuristic rules, which lack the flexibility to cope with fluctuating system states such as changing compute availability, bandwidth variations, or power constraints.
In contrast, the diffusion planning formulates the execution as a conditional generation problem, enabling the system to adaptively explore and refine reasoning paths based on both internal semantics and external feedback.

P2AECF leverages a lightweight generative diffusion model that starts with a high-entropy prior over possible execution sequences. Through iterative denoising, the model converges toward a contextually optimal plan by incorporating prompt-level semantics, inter-agent dependencies, and real-time system metrics. The planner's inputs include the compiled DAG structure, agent metadata (such as latency and energy profiles), and environmental signals (e.g., thermal status, latency bounds, and connectivity quality). Additionally, externally imposed constraints, such as priority levels or energy-saving policies, are incorporated into the planning process.

The output is a ranked set of feasible execution paths, evaluated using multi-objective cost functions such as end-to-end latency, cumulative energy consumption, confidence scores, or task-specific utility. A plan selection module identifies the best candidate execution path, optionally applying ensemble strategies or uncertainty-aware sampling to improve robustness and ensure stability under changing conditions.

One of the distinct strengths of this approach is its continual learning capability. During and after execution, the system collects telemetry data, including agent stalls, bottleneck patterns, and failure cases. This data is fed back into the denoising process, enabling the model to update its internal sampling priors, improving future inference decisions and adapting to long-term changes in dynamic operational environments.
In addition to this continual learning, the diffusion-based planner is trained offline, with only the sampling process running on edge nodes. It uses low-dimensional task DAG and agent metadata, applying a small number of denoising iterations to generate a few candidate plans.

\subsection{End-to-End Execution Workflow}
The P2AECF operates as a tightly integrated, closed-loop pipeline for dynamic task execution in resource-constrained environments. This end-to-end workflow spans from semantic-level task specification to adaptive runtime reasoning, ensuring that edge systems efficiently respond to diverse operational demands.

The process starts when a semantic prompt is issued by a user, sensor, or higher-level controller. This prompt, which can be in natural language or structured format, is processed by the prompt-defined cognition module. The system parses the prompt, extracts its intent, and decomposes the task into an abstract DAG, representing interdependent cognitive operations.
Next, the agent-based modular execution mechanism dynamically assigns each subtask in the DAG to a suitable cognitive agent. This selection process considers metadata descriptors (e.g., input-output formats, latency profiles, accuracy) and real-time system conditions, such as compute availability, battery level, and agent performance. The orchestrator binds agents to DAG nodes based on this information and prepares them for execution.
Once the task graph is instantiated, the diffusion-based inference planner generates a set of candidate execution strategies. These strategies are refined through runtime feedback and heuristics from prior executions, taking into account environmental dynamics and system constraints. The planner selects the most appropriate execution path, which could be sequential, parallel, or hybrid depending on the task graph's topology and QoS objectives.

The agents execute according to the generated plan. During execution, the system continuously monitors telemetry data, including agent-level latency, energy consumption, inference confidence, and communication delay. This telemetry is aggregated and stored for downstream optimization.
Crucially, the final step in the cycle feeds execution metrics back into the system, updating the prompt parsing strategies, agent selection policies, and diffusion planning parameters. This feedback enables continuous system learning and adaptation, optimizing the execution strategy over time.

\begin{figure*}[!t]
\centering
\includegraphics[width=16cm]{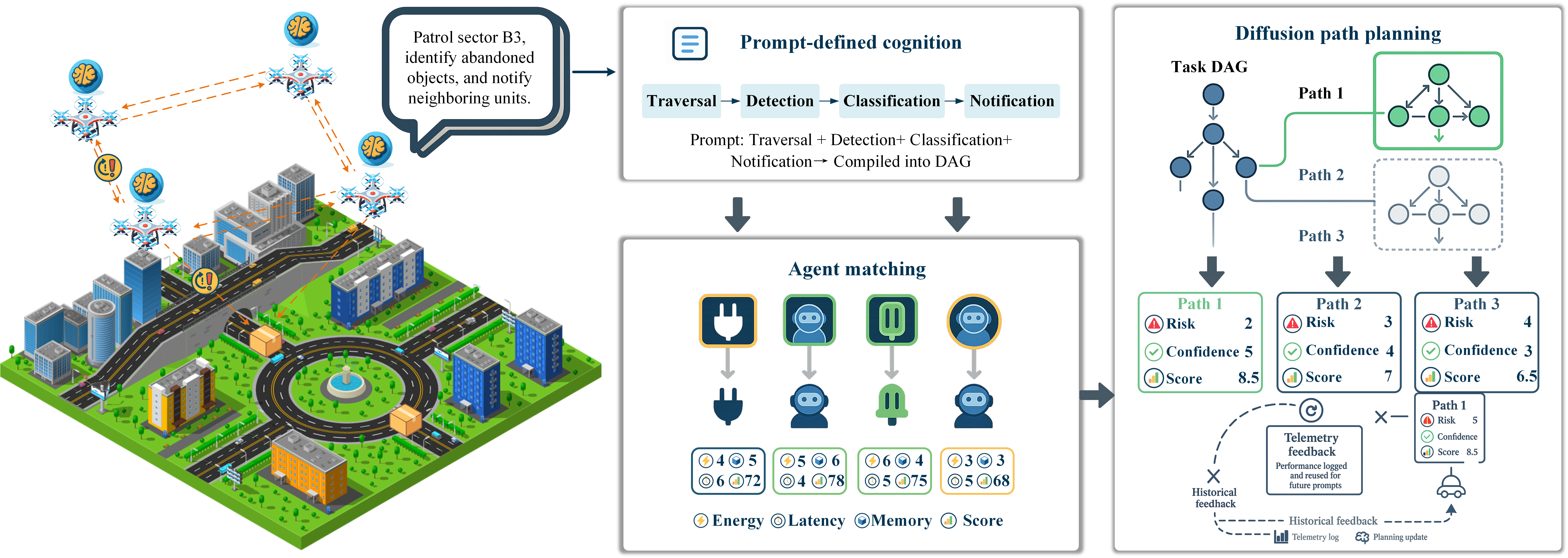}
\caption{Prompt-driven AAV task execution, starting from prompt parsing and DAG compilation, followed by agent matching, diffusion planning, and concluding with telemetry feedback.}
\label{f4}
\end{figure*}

\section{P2AECF-Enabled LAIN} \label{sX}

\subsection{Cognitive Demands in LAIN}
LAINs integrate aerial platforms with terrestrial nodes to enable distributed and real-time cognitive services in mission-critical scenarios such as urban surveillance, disaster response, and intelligent transportation. 
LAINs impose stringent cognitive demands, including ultra-low-latency decision making for tasks like object detection and inter-agent areas where centralized cloud solutions often fall short.
Additionally, these networks operate in dynamic and unpredictable environments characterized by evolving missions, heterogeneous sensor inputs, variable weather conditions, and changes in network topology due to the platform mobility. Given resource constraints on aerial platforms, such as limited computational capacity, energy reserves, and communication bandwidth, cognitive solutions must be lightweight, adaptive, and energy efficient. Furthermore, the effective LAIN operation requires semantic reasoning and an abstract understanding of mission intent, environmental dynamics, and interactions among nodes.

\subsection{Integration of P2AECF into LAIN}
P2AECF aligns well with the operational characteristics of LAINs due to its modular structure, lightweight runtime, and context-aware adaptability. 
LAINs require real-time cognition across distributed aerial and terrestrial nodes, operating under fluctuating connectivity, constrained local resources, and highly variable mission objectives.
P2AECF addresses these challenges by integrating three core components: prompt-defined cognition, agent-based modular execution, and diffusion-based inference planning into the air-ground architecture of LAIN.

The prompt-defined cognition module serves as the semantic interface between users, onboard sensors, and autonomous control agents. In practical deployments, prompts originate from various sources such as ground control centers, AAV sensor feeds, or inter-vehicle coordination protocols. These prompts, either in natural language or structured formats (e.g., JSON templates), describe tasks at a high level of abstraction. The system parses each prompt to extract its core intent and transforms it into an abstract DAG that represents interdependent cognitive operations, without assigning them to specific models. This abstraction enables the system to be flexible and adaptive, allowing it to handle various tasks with different requirements across heterogeneous agents in the LAIN.

The agent-based execution layer connects abstract tasks with executable operations. Each LAIN node, whether an AAV or a terrestrial edge unit, maintains a repository of cognitive agents responsible for specific tasks such as object tracking, anomaly detection, and semantic classification. These agents are selected dynamically at runtime based on metadata descriptors, which include supported input formats, inference latency, energy cost, and historical performance. This selection process ensures that tasks are assigned to the most suitable agent based on real-time resource availability, allowing for efficient task execution and workload distribution across heterogeneous nodes in the LAIN. The modular nature of this layer enables scalability as new agents and nodes can be added to the network without disrupting system performance.

The diffusion-based inference planner generates adaptive reasoning paths by sampling candidate execution plans in response to real-time system states. It continuously monitors contextual signals such as processing load, battery levels, communication status, and thermal conditions, dynamically adjusting the task schedule based on the available resources. This adaptive planning ensures that task execution can be optimized in real-time to balance latency, energy consumption, and task completion rates, making it especially suitable for dynamic environments where network conditions and resource availability can change rapidly. By selecting the most context-appropriate execution path, the planner improves the robustness and responsiveness of the system in highly variable mission environments.

In larger-scale deployments, such as those involving thousands of AAVs or edge devices, P2AECF maintains scalability by breaking down complex tasks into smaller subtasks and distributing them across a hierarchical network of agents. This decentralized approach ensures load balancing and resilient task execution, even in the presence of fluctuating network conditions and resource constraints.

\begin{figure}[!t]
\centering
\subfloat[Impact of network latency on task completion rate.]{
  \includegraphics[width=0.8\linewidth]{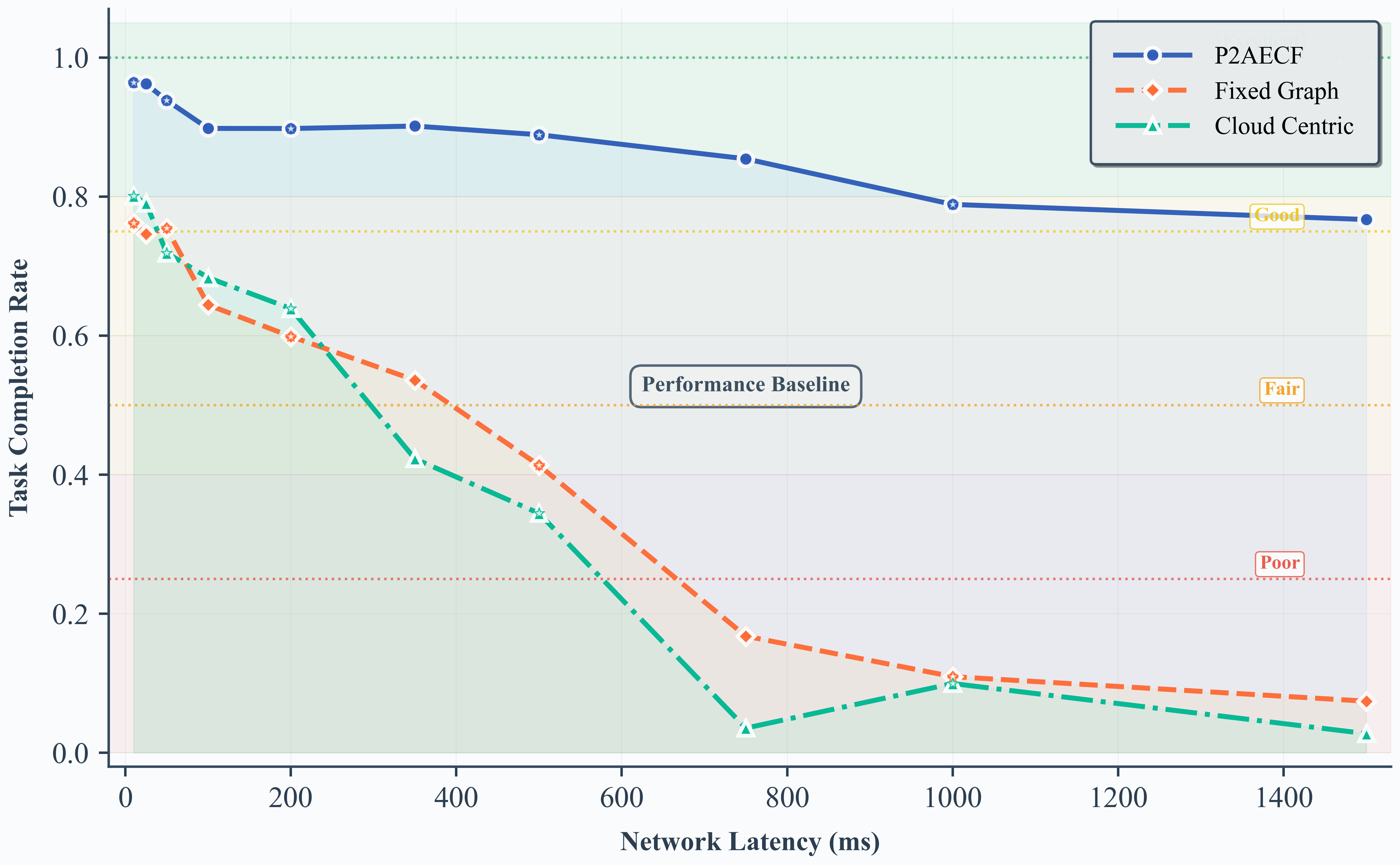}
  \label{f5}
}
\hfill
\subfloat[Comprehensive performance comparison.]{
  \includegraphics[width=0.8\linewidth]{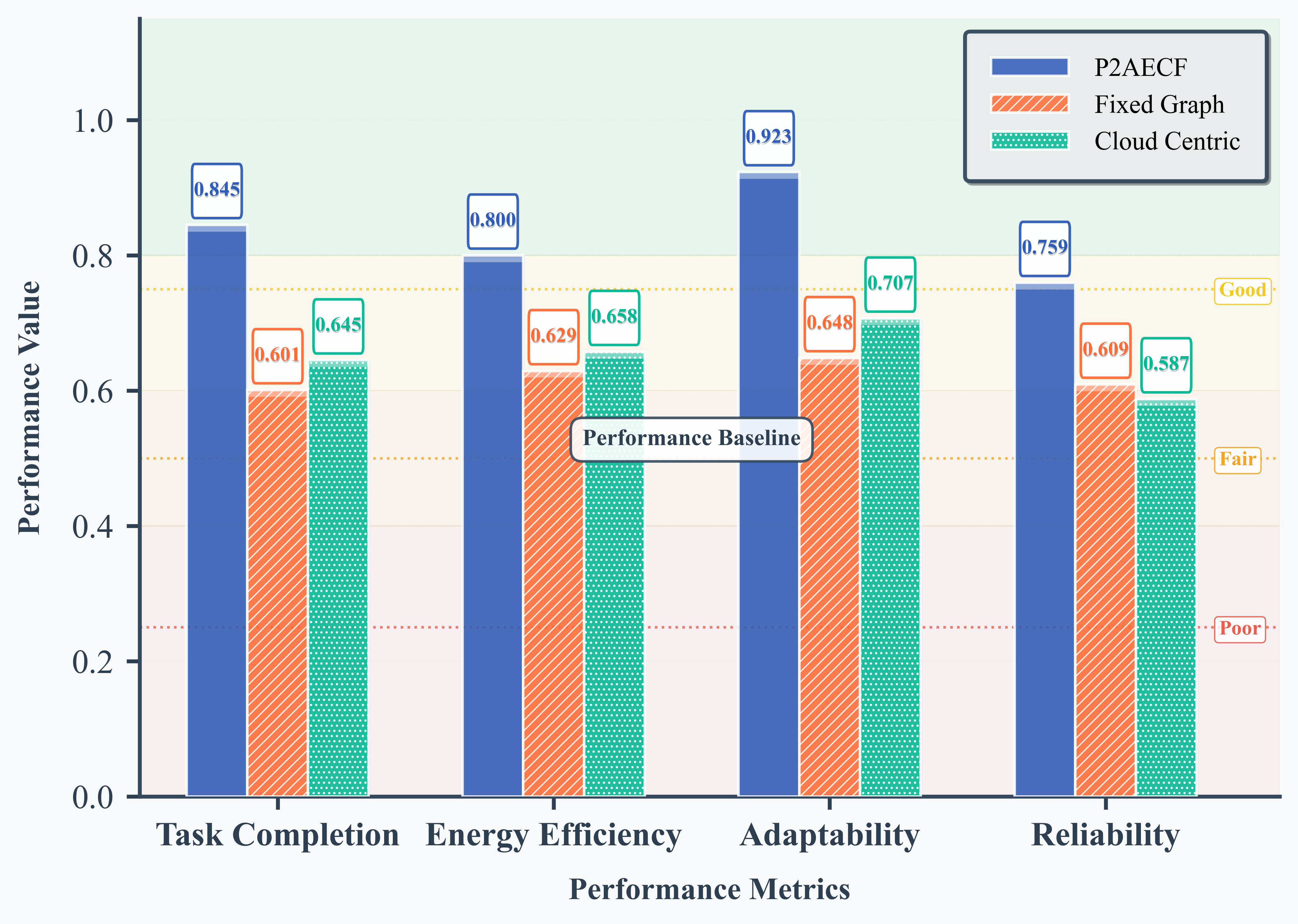}
  \label{f6}
}
\caption{P2AECF performance analysis.}
\label{fig:performance}
\end{figure}

\subsection{Case Study: Coordinated Reasoning for AAV}
Fig. \ref{f4} illustrates a scenario where P2AECF is used for coordinated AAV operations in dynamic missions like patrol or search-and-rescue, involving tasks such as anomaly detection, object tracking, and inter-agent coordination. P2AECF parses semantic prompts into task graphs, dynamically assigning cognitive agents based on real-time resource constraints. The diffusion-based planner adjusts execution plans based on factors like battery level and agent reliability, ensuring mission continuity.
We simulate P2AECF in a discrete-event simulation over a 2km \(\times\) 2km urban grid, with AAVs having 600m communication range, 100 Mbps bandwidth, 10 GFLOPS compute, and 2,000J energy budget. The simulation compares P2AECF with Fixed-Graph (a static pipeline) and Cloud-Centric (remote offloading with intermittent connectivity). Key metrics include task completion rate, energy efficiency, adaptability, and reliability, evaluating the framework's effectiveness, resource utilization, system adaptability, and robustness.

Then, we evaluate the P2AECF framework against the Fixed-Graph and Cloud-Centric baselines under varying latency conditions and key performance metrics. As shown in Fig. \ref{fig:performance}, P2AECF consistently outperforms both. It achieves a task completion rate of 96.4\% under ideal latency (10 ms) and maintains 76.7\% even at 1,500 ms, while the other methods degrade sharply. This robustness results from its dynamic agent assignment and context-aware inference planning. In a broader comparison, P2AECF shows higher scores in task completion (0.845), energy efficiency (0.800), adaptability (0.923), and reliability (0.759). These improvements stem from its modular agent orchestration, closed-loop feedback, and the ability to reconfigure execution paths based on runtime constraints. The results validate P2AECF as a resilient and adaptive edge cognition framework suitable for mission-critical AAV scenarios.

\section{Conclusion}\label{s6}
The article presents P2AECF, a unified edge cognition architecture designed to address the limitations of deploying LAMs in LAINs. By integrating the prompt-defined cognition, agent-based modular execution, and diffusion-controlled inference planning, P2AECF enables edge systems to interpret high-level task intent, compose modular reasoning workflows, and adapt execution strategies based on real-time conditions and resource availability.
The framework effectively decouples cognitive task specification from model implementation, supports fine-grained agent reuse across heterogeneous nodes, and leverages generative planning to optimize execution under uncertainty. Through a representative case study involving collaborative autonomous aerial vehicle operations, we validate ability to achieve low-latency, adaptive, and resilient decision-making in dynamic and resource-constrained environments.
Overall, P2AECF offers a scalable, interpretable, and self-adaptive approach to cognition-native edge intelligence, making it well-suited for future LAIN deployments that require high flexibility and resilience in the complex and dynamic environments.

\bibliographystyle{IEEEtran}
\bibliography{references}
\end{document}